# *Resolving the Missing Financial Data Crisis: A Generative AI Pipeline for SEC 10-K Extraction*

Prisha Nair
*Data Science Program*
*Massachusetts Academy of Math and Science*
Worcester, Massachusetts
ppnair@wpi.edu

Dr. Roee Shraga
*Data Science Program*
*Worcester Polytechnic Institute*
Worcester, Massachusetts
rshraga@wpi.edu

***Abstract*—SEC 10-K filings contain substantial financial information that is not consistently captured in structured datasets, creating a missing-data problem affecting over 70% of firms and half of total market capitalization. This can disproportionately bias quantitative analysis against smaller firms, which may be excluded due to limited available data. Traditional financial extraction methods such as Regular Expressions (Regex) and BERT have been widely used. However, they are highly brittle when parsing complex SEC 10-K filings, which leads to data that is existent in the files being lost since these methods don't consider that a data attribute could be located in a different section or a footnote. This study evaluates several Large Language Models (LLMs), including Llama-3 8B, Qwen-2.5 14B, and Llama-3.3 70B, to figure out individual model strengths and weaknesses when extracting specific attributes from SEC 10-K text. The extraction quality was evaluated across four financial variables of varying structural complexity: Cash and Cash Equivalents (tabular), Short-Term Debt (hybrid), Credit Facilities (narrative), and Research and Development (hybrid). Results show that while smaller models like Llama-3 8B experience performance degradation under complex negative prompting, aligning parameter scale with document complexity yields high zero-shot accuracy. Qwen-2.5 14B excels as a tabular specialist with an 83.33% F1 score on Cash, whereas Llama-3.3 70B effectively navigates dense narrative footnotes, achieving a 76.92% F1 score on R&D. This scalable framework addresses critical information gaps in quantitative finance datasets and eliminates missing-data bias through a more thorough analysis of the SEC 10-K files.**

## I. Introduction

Missing financial data is an important issue in quantitative finance, affecting more than 70% of firms and accounting for approximately half of the total market capitalization [1]. When researchers and analysts attempt to handle these gaps, traditional methods such as dropping firms with missing data introduce significant bias. This effectively disadvantages smaller firms or startups that lack the funds to format documents for regular information extractors that ignore surrounding context [1]. Consequently, even though their data might exist in non-traditional areas of an SEC 10-K filing—a comprehensive annual report submitted to the U.S. Securities and Exchange Commission detailing a company's financial performance—it can be marked as 'NOT_FOUND' discouraging investors due to a lack of information. Extracting accurate, structured attribute-value pairs from complex unstructured financial texts like SEC 10-K filings can prove difficult because companies use highly variable formatting, inconsistent table layouts, and custom terminology across filings, which breaks rigid rule-based parsers [2]. But, it is still necessary to populate and create robust and fair financial databases.

Historically, data extraction has relied on rules-based programming methods or traditional pretrained language models (PLMs) like BERT [3]. However, recent research demonstrates that Large Language Models (LLMs) can perform attribute value extraction with higher overall accuracy and better generalization that specialized BERT-based systems, even with less training data [3]. Effectively leveraging LLMs requires strategic prompt engineering, such as processing input data into strict JSON formats and breaking down tasks into highly specific, targeted queries [4].

To optimize LLM extraction, prompts must provide unambiguous definitions of target metrics and explicitly outline what the model should not do [5]. Furthermore, separating instructions from the raw data using distinct tags and forcing casual reasoning by requiring the model to write a step-by-step explanation before outputting its final classification, drastically improves performance as well [6].

Providing clear negative constraints is a standard practice in prompt engineering to give the LLM strict boundaries to follow [5]. However, applying complex exclusionary rules to smaller parameter LLMs within dense financial contexts reveals a significant limitation. During the extraction of nuanced metrics like Research and Development expenses or Short-Term Debt, introducing specific negative constraints (e.g., instructing the model not to extract tax credits) causes smaller, 8B parameter models to suffer from constraint paralysis—a phenomenon where dense or overlapping rules can over-restrict a model's processing pathways and cause its attention mechanism to fracture. Parsing fine-grained negative prompts causes smaller models to experience dramatic performance degradation as contextual rules conflict [7]. Instead of correcting edge cases, the model's attention mechanism fractures, leading to failures on the core extraction tasks it was meant to succeed with.

To address this gap, this paper benchmarks traditional rules-based regular expressions (Regex) against a spectrum of open-weight LLMs, including Llama-3 8B [8], Qwen-2.5 14B [9], and Llama-3 70B [10]. Rather than relying on existing benchmarks, this paper constructs and hand-annotates new benchmarks from scratch, as shown in Table II. These benchmarks are built using text chunks and variable info extracted from the FinReflectKG benchmark—an existing knowledge graph frequently used for testing information extraction from SEC 10-K files using LLMs [11]. This is due to the fact that although some benchmarks exist for SEC 10-K files, many are text heavy, like FinReflectKG, and require further enhancement to be used specifically for numerical information extraction. By evaluating the LLMs across four distinct financial structures (Cash and Cash Equivalents, Short-Term Debt, Credit Facilities, and Research and Development), this study demonstrates that the required parameter scale of an LLM is directly dictated by the structural complexity of the document. Specifically, this paper makes three key contributions: (1) it constructs and hand-annotates a novel numerical extraction benchmark; (2) it designs an open-source extraction pipeline; (3) it provides an empirical analysis demonstrating how document complexity dictates required model scale to find balance between rules-based prompts and parameter counts in LLMs.

## II. Data Preparation

Benchmarking data extraction from unstructured financial documents requires a reliable system that can isolate specific numbers from messy text. This framework connects raw SEC filings to organized data by establishing consistent labeling rules, basic rule-based tests, and scalable AI models to measure how well different systems handle document complexity.

### *A. Target Variable Selection and Initial Data Collection*

In order to effectively test the LLM extraction viability for numeric data from unstructured SEC 10-K filings, there needed to be a benchmark dataset. Many financial datasets are not numerically heavy, rather they have more textual data such as large text chunks and variable descriptions as well as ticker values [11]. For this paper, it was necessary to create a benchmark that isolated the important numbers from the text chunks provided using guidance from the variable names. Additionally, in order to give the number some context, there needed to be a corresponding year, a numeric value isolated from the text chunk, as well. To construct an optimal ground-truth dataset, we adapt FinReflectKG [11]. While the knowledge graph provides a repository containing over 10 different numerical attribute types, many of these attributes are sparse or lack the structural variety needed to evaluate modern AI models. To address this, we isolate four core variables to create hand-annotated benchmarks, summarized in Table II. They were specifically chosen for their universal nature and ability, which is summarized in Table I, to test distinct aspects of model extraction behavior across varying document complexities.

#### *1) Credit Facilities*

Credit Facilities represents the low-structure, narrative-dense text chunks. Embedded in legal prose, this variable tests localized sentence parsing. It requires isolating active borrowing amounts while actively avoiding undrawn accordion features, floating interest rate margins, and administrative fee clauses. FinReflectKG contains 52,718 rows from which 100 rows were sampled [12].

#### *2) Short Term Debt*

Short Term Debt represents a hybrid structure where it presents itself in both debt maturity tables and descriptive narrative notes. It evaluates whether models can separate current liabilities from long-term totals, ignore facility limits, and spot implicit zero balances (e.g., text chunk's asserting "no commercial paper outstanding"). FinReflectKG yielded 37,637 rows from which 100 rows were randomly sampled [12].

#### *3) Cash and Cash Equivalents (CHE)*

This attribute serves as the high-structure, tabular baseline. Its primary format is rigid Balance Sheet tables, therefore CHE tests a model's ability to map column headers, which are mostly fiscal years, directly to grid values. Querying FinReflectKG resulted in only 47 rows that had any connection to this attribute [12]. This was justified since CHE serves as a rigid, highly populated tabular benchmark, whereas other variables suffer from severe missing data rates across Compustat [1].

#### *4) Research and Development (R&D)*

R&D acts as a complex hybrid metric frequently split across Income Statement lines and footnoted disclosures. R&D was chosen because it introduces heavy contextual distractors, including forward-looking spending projections ("we expect to invest"), revenue percentage ratios ("R&D represented 12% of net sales), and one-time in-process R&D acquisition write-offs. FinReflectKG returned 16,843 rows, from which 100 rows were sampled to form the final ground truth dataset [12].

TABLE I. Variable Composition and Scarcity Context

| **Financial Variable** | **Rows** | **Structural Complexity** | **Scarcity/Extraction Difficulty** |
|---|---|---|---|
| Cash and Cash Equivalents (CHE) | 47 | Tabular (Low) | Used as a highly populated, rigid benchmark variable (low scarcity). |
| Short Term Debt | 100 | Hybrid (Medium) | Frequently hidden within 'Total Debt' totals; moderate missing data rates. |
| Credit Facilities | 100 | Narrative (High) | Severe scarcity; often buried in dense legal paragraphs without structured tables. |
| Research and Development (R&D) | 100 | Hybrid (High) | High scarcity; highly susceptible to ratio distractors and future estimates. |

### *B. Ground Truth Dataset Construction*

To establish a reliable benchmark for evaluating model precision, recall, and F1 scores, standardized ground truth datasets were constructed for each variable. Using the samples selected from Cash and Cash Equivalents, Credit Facilities, Research & Development, and Short-Term Debt, (The sample sizes can be seen in Table I) each row had its own chunk of text which was manually annotated to establish values for Fiscal Year, a Consolidated Flag, and the Ground Truth Float.

TABLE II.

TABLE III. GROUND TRUTH BENCHMARK AND FIELD DESCRIPTIONS

| **Field Name** | **Data Type** | **Description & Extraction Criteria** |
|---|---|---|
| Fiscal_Year | Integer | The target reporting period (e.g., 2021, 2022, 2023) directly tied to the financial observation within the 10-K text chunk. |
| Consolidated_Flag | Boolean | Indicates whether the value reflects total consolidated entity figures (TRUE) or localized subsidiary, segment, or non-consolidated breakdowns (FALSE). |
| Ground_Truth_Float | Float | The exact quantitative monetary value extracted from the filing segment, isolating historical actuals while dodging forward-looking estimates and ratio distractors. |

Strict variable-specific ontology rules governed the manual annotation process and would also later guide the prompt engineering process. For CHE, only primary unrestricted balance sheet lines were annotated, while ambiguous pension fund assets (e.g., percentage of plan assets) and historical beginning-of-period cash flow balances were classified as 'NOT_FOUND'. For Credit Facilities, active borrowing capacities were isolated while undrawn accordion features and floating interest rate margins were ignored. For Short-Term Debt, current liabilities were differentiated from long-term debt totals, and zero balances were annotated when text chunks said, "no commercial paper outstanding". Finally, for R&D, annotations isolated historical operating expenses, explicitly rejecting future projected expenditures and one-time in-process R&D acquisition charges.

## III. EVALUATION FRAMEWORK

### A. Rule-Based Regular Expression (Regex) Pipeline

To demonstrate the limitations of rules-based programming, Regular Expression (Regex) engines served as a primary baseline. The Regex scripts shared uniform features, including scale detectors for phrases like 'in millions' to correctly scale numbers into full-length floats. Additionally, they shared an absolute consistency enforcement rule which mandated that if the Fiscal Year was 'NOT_FOUND', the Float and Consolidated Flag must also evaluate to the same. This is because a number without a year and a year without a number both lack enough context to be useful, therefore the whole row is marked as 'NOT_FOUND'.

Beyond these shared features, the pipelines were tailored to the structural characteristics of each variable. The CHE script utilized continuous string anchoring and a 50-character backwards context window to reject traps such as "restricted" or "beginning," alongside a "year-dodging" failsafe to prevent the extraction of 4-digit column headers. Conversely, because Credit Facilities reside in dense paragraphs, its respective script utilized a narrative-focused design that dynamically split text into discrete sentences. This engine relied on sentence-level anchors and strict rejection clauses for terms like "accordion," or "interest rate," while incorporating local word-scale multipliers (e.g., "$2.5 billion"). For the hybrid variables (Short-Term Debt and R&D), the engines split segments by both newlines and periods to process tables and paragraphs simultaneously. The Short-Term Debt engine introduced explicit zero-detection rules for phrases like "none outstanding," while the R&D engine deployed aggressive filters to reject forward-looking statements ("expect", "plan") and ratio distractors.

### B. Discriminative Pretrained Models (BERT)

Following the Regex evaluation, we evaluated traditional pretrained language models (PLMs) utilizing BERT-based architectures as a discriminative machine learning baseline. While effective for simple token classification, BERT models struggled with the spatial disconnect inherent in flattened SEC tables, failing to accurately correlate header rows containing fiscal years with their respective numerical cells. All in all, the model exhibited zero-shot fragility and failed to generalize across non-standard accounting phrasing, yielding a 0.00% Joint F1 Score during the Cash and Equivalents extractions benchmark. For this reason, the other variables were not tested with BERT, since it was assumed that the results would be similar.

### C. Prompt Engineering Techniques

To address the shortcomings of discriminative models, the study transitioned to LLMs, utilizing the open-source Llama-3 8B model [8] to establish prompt engineering architectures. Specifically, prompts employed Chain-of-Thought (CoT) reasoning to force step-by-step logical explanations prior to outputting final JSON fields, alongside few-shot in-context learning via variable descriptions and examples. Instructions were separated from raw data using explicit tags to give the model precise context for text chunk extraction.

### D. Model Scaling and Cross-Model Validation

To maximize extraction accuracy and evaluate parameter scale, the most effective prompt templates developed during Llama-3 8B iterations, specifically those utilizing CoT reasoning and in-context learning prior to constraint paralysis, were transferred to higher-parameter open-source models: Qwen-2.5 14B and Llama-3.3 70B. Qwen-2.5 14B was evaluated to see how mid-sized tabular architectures handled the transition to unstructured text. Subsequently, Llama-3.3 70B was tested using these optimal prompts to determine how massive parameter scaling improves recall and precision when navigating complex hybrid documents. All models were systematically executed across the four benchmark datasets under identical evaluation conditions to generate cross-model comparisons of precision, recall, and joint F1 scores.

### E. Joint Strict Match F1 Scores

Model extraction performance was evaluated using Joint Strict Match F1, defined as the harmonic mean of precision ($P$) and recall ($R$) across all target fields simultaneously:

$$F1 = 2 \times \frac{P \times R}{P + R}$$

A true positive requires the model to correctly extract the exact Fiscal Year, Consolidated Flag, and Float value concurrently.

## IV. Results

Financial data extraction was evaluated by comparing traditional deterministic engines against modern open-source generative models. To clearly illustrate the impact of both parameter scale and document structure on extraction accuracy, the findings are presented in three distinct phases. This progression begins by establishing the performance floor with baseline methodologies, details the issue of constraint paralysis in smaller LLMs, and culminates with the optimized performance of high-parameter architectures.

### A. The Baseline Methods

The initial evaluation of traditional rules-based programming and discriminative pretrained language models revealed severe limitations when processing unstructured SEC 10-K data. The Regex pipelines struggled to contextualize spatial relationships and accurately dodge text distractors. On Cash and Cash Equivalents (CHE), the Regex baseline achieved a Joint Strict Match F1 Score of only 4.88%. Performance remained similar across the hybrid and narrative datasets, scoring a 15.93% Joint F1 on Credit Facilities and a 5.76% Joint F1 on Short-Term Debt. Most notably, the Regex engine completely failed on the highly complex Research and Development (R&D) dataset, scoring a 0.00% Joint F1 due to its inability to reliably filter out forward-looking estimates and ratio percentages.

Similarly, the BERT discriminative architecture exhibited no flexibility in extracting from the messy SEC 10-K tables. During the CHE extraction benchmark, the BERT model failed to map the correct fiscal year column headers to their corresponding numerical cells in flattened tables, yielding a Joint Strict Match F1 Score of 0.00%.

### B. Generative Models and Constraint Paralysis

Transitioning to generative models, the open-weight Llama-3 8B model established a significantly higher accuracy baseline but revealed a critical vulnerability during iterative prompt engineering. Initial prompt templates yielded moderate success, achieving a Joint F1 of 40.29% on Short-Term Debt, 34.48% on R&D, 62.75% on CHE, and 34.62% on Credit Facilities.

However, as the prompts were engineered to include complex negative constraints to eliminate false positives, such as instructing the model to explicitly ignore R&D tax credits or unused commercial paper capacity, overall extraction accuracy degraded. Specifically, while precision remained artificially high, the 8B model suffered a steep drop in recall as valid extractions were incorrectly filtered out. The 8B parameter model suffered from "constraint paralysis," where it exhibited semantic drift and instruction-following breakdown under rule fatigue.

For example, on Short-Term Debt, adding capacity constraints caused the Joint F1 score to drop from 40.29% to 31.08%. Additionally, on R&D, adding tax credit constraints caused the Joint F1 score to drop from 34.48% to 27.91%. Finally, on Credit Facilities, attempting to filter out accordion features and future maturities dropped the Joint F1 down to 18.95%.

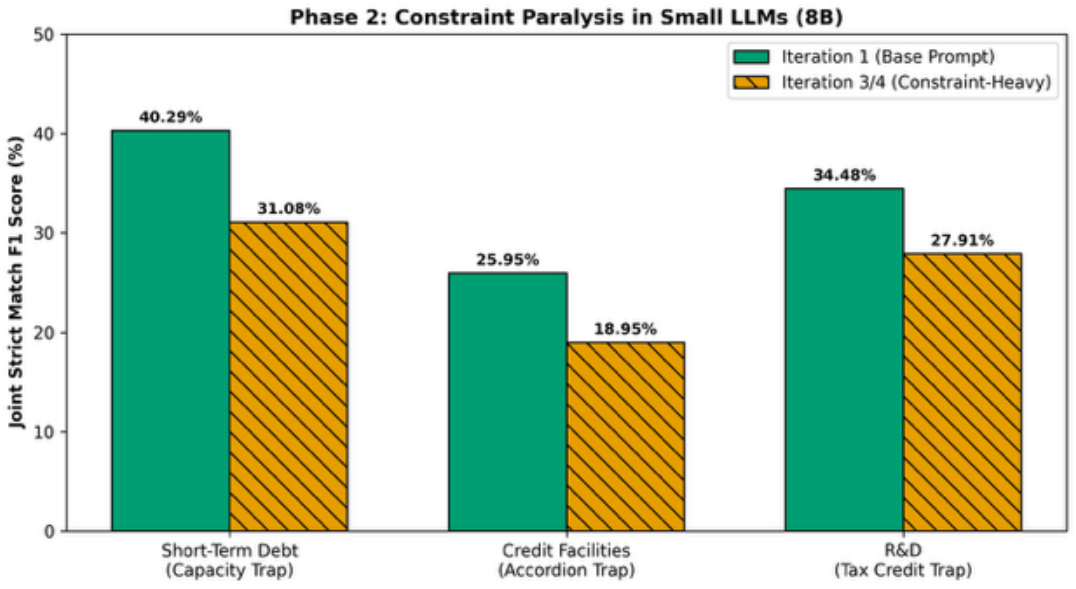

Fig. 1. The onset of constraint paralysis in the 8B-parameter Llama-3 model. The introduction of complex negative constraints (Iteration 3/4) resulted in attention fracturing and a severe degradation of joint extraction accuracy compared to the initial baseline prompt (Iteration 1).

### C. Model Scaling vs. Document Complexity

To maximize accuracy, the optimized prompt templates were transferred to higher-parameter architectures. The Qwen-2.5 14B model demonstrated exceptional proficiency on rigid, tabular data, achieving an 83.33% Joint F1 Score on the Cash and Cash Equivalents dataset. However, the 14B architecture struggled significantly when navigating dense legal paragraphs, scoring only a 27.50% Joint F1 on Credit Facilities and a 34.48% on Short-Term Debt. This indicates that while 14B parameters are sufficient for tabular reasoning, they lack the localized contextual awareness required for narrative extraction.

TABLE IV. Precision, Recall, and Joint Strict F1 Scores Across Best-Performing Model Configurations

| Target Variable | Model Configuration | Precision (%) | Recall (%) | F1 (%) |
|---|---|---|---|---|
| Cash and Cash Equivalents | Regex Pipeline | 12.50 | 3.03 | 4.88 |
| | Llama-3 8B (Base) | 65.20 | 60.40 | 62.75 |
| | **Qwen-2.5 14B** | **86.96** | **80.00** | **83.33** |
| | Llama-3.3 70B | 76.00 | 70.21 | 72.34 |
| Short-Term Debt | Llama-3 8B (Base) | 44.12 | 37.14 | 40.29 |
| | Llama-3 8B (Constrained) | 52.17 | 22.58 | 31.08 |
| | Qwen-2.5 14B | 36.50 | 32.60 | 34.48 |
| | **Llama-3.3 70B** | **51.02** | **46.15** | **48.45** |
| Credit Facilities | Regex Pipeline | 22.00 | 12.40 | 15.93 |
| | Llama-3 8B (Base) | 38.00 | 31.50 | 34.62 |

| | Qwen-2.5 14B | 31.25 | 24.39 | 27.50 |
|---|---|---|---|---|
| | **Llama-3.3 70B** | **46.15** | **38.71** | **42.11** |
| **Target Variable** | **Model Configuration** | **Precision (%)** | **Recall (%)** | **F1 (%)** |
| Research & Development | Regex Pipeline | 0.00 | 0.00 | 0.00 |
| | Llama-3 8B (Base) | 37.93 | 31.43 | 34.48 |
| | Qwen-2.5 14B | 30.00 | 25.00 | 27.42 |
| | **Llama-3.3 70B** | **80.00** | **74.19** | **76.92** |

Ultimately, massive parameter scaling via the Llama-3.3 70B model successfully resolved both tabular and narrative bottlenecks. Utilizing Chain-of-Thought reasoning, the 70B model effectively balanced complex negative constraints without fracturing. It achieved a 72.34% Joint F1 on CHE and a 42.11% Joint F1 on Credit Facilities. Most notably, the 70B model dominated the highly complex Research and Development dataset, effectively ignoring future estimates and ratio distractors to achieve a remarkable 76.92% Joint F1 Score.

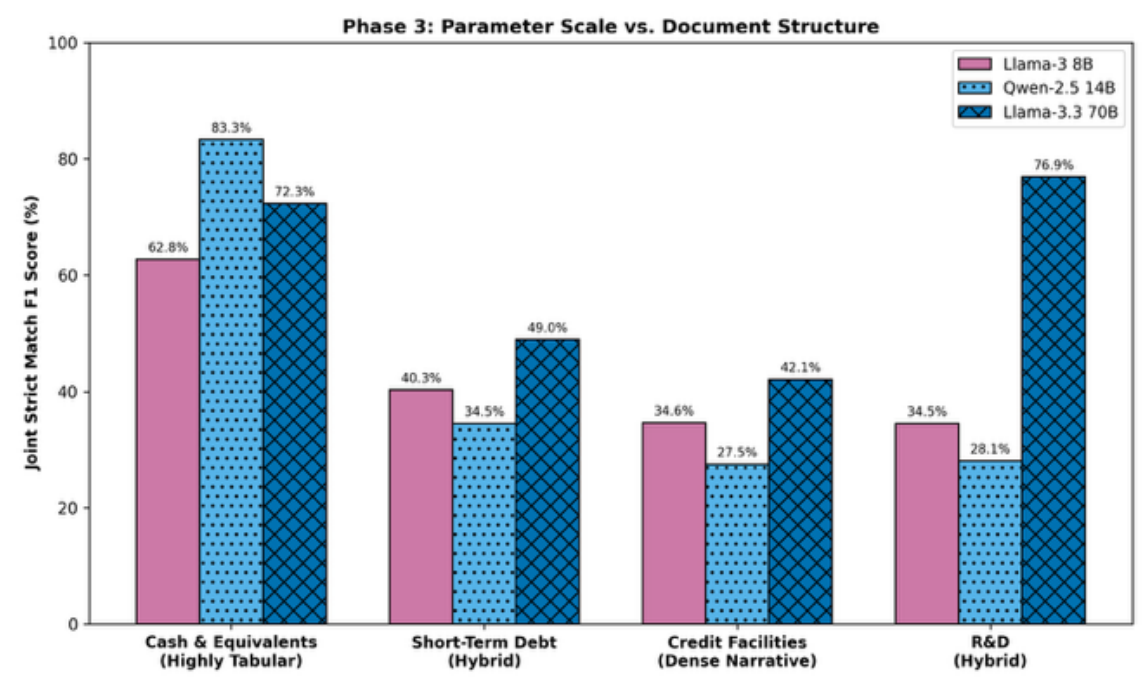

Fig. 2. Cross-model comparison of Joint F1 scores demonstrating the relationship between parameter scale and document complexity. Qwen-2.5 14B excels on tabular variables (Cash & Equivalents), while massive scaling via Llama-3.3 70B is required to successfully parse narrative and hybrid variables (R&D, Credit Facilities).

## V. Discussion & Conclusion

A primary contribution of this study is identifying "constraint paralysis" in small-parameter models, where applying exclusionary rules to the 8B Llama-3 model degraded Joint Strict Match accuracy due to rule fatigue. Cross-model evaluation revealed that optimal zero-shot extraction requires aligning parameter scale with document structure: Qwen-2.5 14B excels as a tabular specialist on structured balance sheets, whereas massive parameter scaling via Llama-3.3 70B resolves contextual bottlenecks to navigate narrative distractors in R&D and Credit Facilities.

Practically, this framework addresses the missing financial data crisis, affecting over 70% of firms and half of total market capitalization, through a local open-source pipeline. Deploying these models locally eliminates per-token API costs (such as the ~$0.03 to $0.05 per 1K tokens incurred by GPT-4), reducing inference overhead to zero outside of hardware costs. Institutions can optimize costs by routing standardized balance sheet tables to fast 14B models while reserving 70B models for deep narrative footnote analysis.

This study is bounded by dataset subsets (47–100 rows per variable from FinReflectKG), single-annotator ground-truth construction, an open-source architectural scope, and single-run deterministic evaluations. Results confirm that deterministic Regex is too brittle for nuanced metrics like R&D, while smaller generative models (8B) suffer from constraint paralysis under complex negative constraints. Overall, achieving optimal extraction accuracy requires aligning parameter scale with physical document structure: 14B models excel as tabular specialists, whereas 70B architectures paired with Chain-of-Thought reasoning are essential for dense narrative footnotes. Future work will expand sample sizes, run scripts multiple times, explore fine-tuning smaller models to eliminate rule fatigue, and integrate this extraction pipeline with statistical imputation methods to resolve the missing data crisis.